\documentclass[]{article}
\usepackage[a4paper, total={6in, 9in}]{geometry}
\usepackage[colorinlistoftodos]{todonotes}
\usepackage{graphicx}\graphicspath{{./images/}}
\usepackage{subfig}
\usepackage{amssymb}
\usepackage{amsopn}
\usepackage{amsmath}
\usepackage{amsfonts}
\usepackage{amsthm} % Math packages\\
\usepackage{epsfig}
\usepackage{epstopdf}
\usepackage{multirow}	
\usepackage{adjustbox}
\usepackage{sectsty} % Allows customizing section commands
\usepackage{setspace}% http://ctan.org/pkg/setspace
\usepackage{xparse}
\usepackage[ruled,vlined]{algorithm2e}
\usepackage{algpseudocode}
\usepackage[absolute]{textpos}
\usepackage{float}
\usepackage{caption}
\usepackage{hyperref}
\usepackage{bookmark}
\usepackage{pgfgantt}

\newcommand{\mnistImagesWidth}{0.15}
\providecommand{\keywords}[1]{~\\\textbf{\textrm{Keywords:}} #1}
\title{Correlated discrete data generation using adversarial training}

\author{%
\begin{tabular}{c} Shreyas Patel$^1$  \\ \\
				   Raj Derasari$^1$  \\ \\
\end{tabular} \and
\begin{tabular}{c} Ashutosh Kakadiya$^1$ \\ \\
				   Rahul Patel$^{1,2}$ \\ \\
\end{tabular} \and
\begin{tabular}{c}  Maitrey Mehta$^1$ \\ \\
				   Ratnik Gandhi$^1$ \\ \\
\end{tabular}  \and  ${}^1$School of Engineering and Applied Science, Ahmedabad University\\~\\  ${}^2$Logistixian Technologies Private Limited}

\begin{document}
%\begin{minipage}{\textwidth}
\maketitle
%\end{minipage}
\begin{abstract}
Generative Adversarial Networks (GAN) have shown great promise in tasks like synthetic image generation, image inpainting, style transfer, and anomaly detection. However, generating discrete data is a challenge. This work presents an adversarial training based correlated discrete data (CDD) generation model. It also details an approach for conditional CDD generation. The results of our approach are presented over two datasets; job-seeking candidates skill set (private dataset) and MNIST (public dataset). From quantitative and qualitative analysis of these results, we show that our model performs better as it leverages inherent correlation in the data, than an existing model that overlooks correlation.\\
\end{abstract}
\section{Introduction}\label{section:Introduction}
Recent advances in computational power have offered a significant boost to learning techniques, fueling applications of Artificial Intelligence(AI) in varied domains -- disease diagnoses \cite{H92}, predicting bank failures \cite{DJ09} and evaluating faults in sports movements \cite{B06}, etc.
Another boost has come from advances due to novel architectures and training techniques: Deep Neural Networks (DNNs), Convolutional Neural Networks (CNNs), Generative Models, Autoencoders amongst many.

\textrm{One key requirement for supervised learning technique in AI is availability of sufficiently big, labeled, and clean dataset. The availability of such dataset, along with the training technique and computing power, form the basis of an effective AI agent. However, in many real-world scenarios, there is a lack of task-specific datasets (e.g. due to privacy,  access restrictions, licenses, copyright, patents and access fees, some of the following datasets are hard to find - medical health-care, criminal background, state policies, employee profiles) or the available data may not be clean (e.g. missing values or corrupted values). Under such scenarios, building human-level AI agents (models) is difficult.}  

Recent papers on generative models\footnote{Generative models \cite{KMR14} are a semi-supervised approach meant for developing models that provide an effective generalization to a large unlabeled data set on the basis of a small labeled dataset. They find use in wide ranges of applications like image search, genomics and natural language parsing, where data is largely unannotated.} attempt to address one of these issues of lack of data. 
Generative Adversarial Networks (GANs) \cite{GPM14}, a type of generative model, help us mitigate the issue of data unavailability by synthesizing new realistic samples. Following this work, there have been several attempts in the direction of generating image/continuous data \cite{GPM14,SPT16}. Efforts have also been made to make these synthetic samples look more realistic \cite{SPT16}. 
A similar set of early efforts towards generating text/discrete dataset has been reported in \cite{CLZ17,YZW17,ZKZ17,SRD17}. These efforts can be sub categorized into generating text tokens \cite{CLZ17} and generating realistic natural language \cite{YZW17,ZKZ17,SRD17}. It must be noted here that the latter problem is harder as it requires generative models to capture semantic structures of underlying sentences \cite{Ian16}.
In this paper, our focus is on addressing the former problem, i.e., of generating text tokens/discrete data.

We approach this problem considering a scenario in which our aim is to build a text based career counseling AI agent in the domain of employment related services. The most basic ability required by such an agent is as follows:
\begin{itemize}
\item Suggest the best profession to an individual based on their existing skills.
\item Suggest a new set of skills to acquire, considering career preferences and existing skills.
\end{itemize}

If we want to accomplish these tasks using an AI agent, then the problem essentially translates to conditional discrete data generation. In other words our aim is to generate skill set conditioned on either profession, or existing skills, or both. However, in absence of open-source candidate profile data, building such models (AI agents) for career progression counseling is difficult. In this case, we can resort to some models such as generative models (specifically, Generative Adversarial Networks (GANs)\cite{GPM14}) or Variational Autoencoder (VAE)\cite{KW13} that can help generate the required synthetic data.

%In absence of open-source candidate profile data building models (AI agents) for such career progression counselling is difficult.
%In this work, we model skills from a candidate's profile as a binary vector (similar to the model presented in \cite{CBM17}) and with the help of GANs show how to generate synthetic skills. An important difference between our work and results reported in \cite{CBM17} is that our model plays the minimax optimization game over conditional input ($D(x | y)$) vs. \cite{CBM17} optimizing only over input data ($D(x)$), \textbf{where $D(\cdot)$ is discriminator output}. 
%In this case we can resort to some models such as generative models (specifically, Generative adversarial networks (GANs)) or Variational Autoencoder (VAE)\cite{KW13} that can help generate the required synthetic data. It must be noted that even for training these models we need relatively small dataset of candidate profiles. VAEs formulate the problem in the framework of probabilistic graphical models. It generates synthetic data by maximizing the lower bound on the log likelihood of the input datapoints.	\\

In this work, we model skills from a candidate's profile as a binary vector (similar to the model presented in \cite{CBM17}), and with the help of GANs, generate synthetic skills.
An important difference between our work and \cite{CBM17} is that our model plays the minimax optimization game over conditional input ($D(x | y)$) vs. \cite{CBM17} optimizing only over input ($D(x)$), \textrm{where $D(\cdot)$ is discriminator output, $x$ is a skill vector, and $y$ is a profession vector}.
Also, we use Correlational Neural Network \cite{SMHB15} to better capture the relation between a profession and skill set, and among skills of a particular skill set.
It must be noted that generating realistic sentences of a language is a difficult task, especially due to grammar and semantics of a sentence. This not only focuses on the right set of words but also the right order and context.
In this work, we focus on the problem of generating correlated discrete data without worrying about the order between them.

Recently, a combination of Recurrent Neural Netowrks (RNNs) and Deep Q Networks \cite{DBLP:journals/corr/Guo15b} has been used to generate text that use Deep Reinforcement Learning \cite{DBLP:journals/corr/abs-1708-05866}.
The method uses sequence to sequence learning, mainly applied in machine translation, text rephrasing and question answering.
Further, the approach based on Policy Networks\footnote{For more information, please refer \url{http://karpathy.github.io/2016/05/31/rl/.}} for generating text uses the idea of mapping a state in a game to the probability of occurrence of each action given the state. The player and its opponent(adversary) both are trained using this network. As used in seqGAN \cite{YZW17} and maskGAN \cite{fedus2018maskgan}, policy networks first applies a policy algorithm (such as Monte Carlo search in seqGAN or an actor-critic model in maskGAN) and then offers rewards to players for optimization (instead of penalties). 
This approach is not very effective when purpose is to generate discrete text tokens - that do not appreciate positional context as set of tokens in a sentence do.
For example, in seqGAN a sequence of words is a state and the predicted next word is an action. After the new word prediction, the new sequence forms a new state and then again a new word is predicted.  In our work, an output is a prediction of discrete data that does not have sequence or positional context. For these reasons, methods based on Deep Q Networks and Policy Networks are not relevant to the problem addressed in this paper. 

To the best of our knowledge, this is amongst the early attempts to preserve correlation in generated data and is scalable to datasets which are discrete. The rest of this paper is organized as follows: Section \ref*{section:Model} presents our model and algorithms for generating conditional skills for an input candidate profile.  Appendix \ref*{appendix:datapreprocessing} presents challenges for generating initial data for training our generative model while subsection \ref*{subsection:ExpAndResults:Results} details our experiments. We present our concluding remarks in section \ref*{section:FutureWork} with potential line of future work.
%%%%%%%%%%%%%%%%%%%%%%%%%%%%%%%%%%%%%%%%%%%%%%%%%%%%%
%%%%%%%%%%%%%%%%%%%%%% Model %%%%%%%%%%%%%%%%%%%%%%%%
%%%%%%%%%%%%%%%%%%%%%%%%%%%%%%%%%%%%%%%%%%%%%%%%%%%%%
\section{CorrGAN Model}\label{section:Model}
  In this section, we present our model and an algorithm for generating correlated discrete (binary) data using adversarial training. The model comprises of primarily two parts; Correlational Neural Network(CorrNN) \cite{SMHB15} and Generative Adversarial Network (GAN) \cite{GPM14}. GAN face difficulty in generating discrete data as it is difficult to pass the gradient update from discriminator $D$ to generator $G$. To mitigate this challenge, we make use of CorrNN, that essentially learns to project discrete data into some continuous latent space.

CorrNN is an Autoencoder based approach for Common Representation Learning. It takes into consideration the correlation in input data while regenerating the same. After training CorrNN, its encoder network ($Enc$) learns to map discrete input space $\mathbb{Z}^{|c|}_{+}$ to continuous latent space $\mathbb{R}^{h}$ and decoder network ($Dec$) learns to reconstruct the input space from latent space, which is continuous. 

Next, we train the GAN, where $G$ learns to map random seed to continuous latent space. The dimensionality of this latent space is same as that of the latent space in CorrNN. Hence, the output produced by the $G$, acts as an input to the Decoder network of CorrNN.
%The Decoder network then outputs a \textit{continous vector} %\textbf{discrete vector after thresholding
%(it is not required here)} is passed as a fake input to train $D$. Simultaneously, $D$ is also shown real/input discrete vectors, which helps it to discriminate between real and fake. The gradients are then passed back from $D$ to Decoder of CorrNN and from Decoder to G.\\
% The Decoder network then outputs a discrete vector (initially continuous, discretized after appropriate thresholding)
The Decoder network then outputs a vector which is passed as a synthetic input to train $D$.
Simultaneously, $D$ is also shown real discrete vectors from the input data, which helps it to learn discriminating whether an input sample is real or synthetic. The gradients are then passed back from $D$ to Decoder of CorrNN and from Decoder to $G$.
\begin{figure}[H]
	\centering
	\includegraphics[scale=0.5]{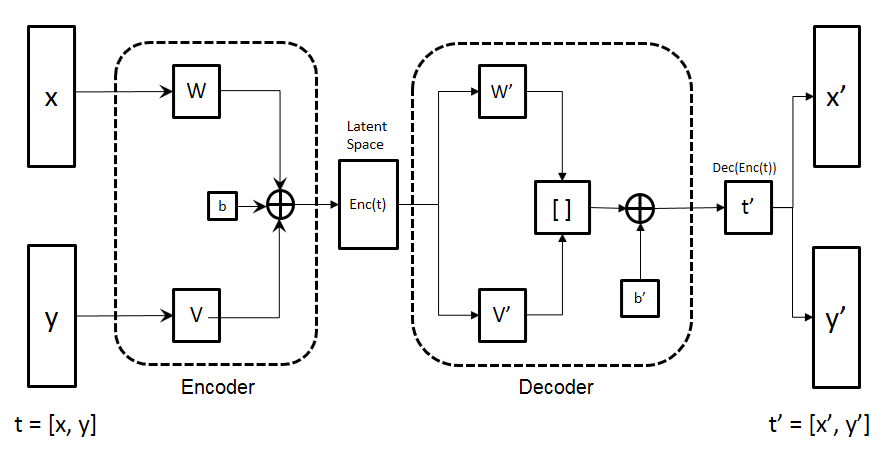}
	\caption{Block diagram of Correlational Neural Network}
	\label{fig:autoencoder}
\end{figure}

Let $T$ be a set of $n$ binary correlated vectors $T = \{ t_{i} \mid 1 < i < n  \}, i \in \mathbb{Z}_{+} $. Input $t_{i}$ is divided into two mutually exclusive subsets such that,
\begin{align}
&t_{i} = [x_i, y_i].
\label{equation:t_x_y}
\end{align}
The Encoder network takes $x_i$ and $y_i$ as an input and map it to a continuous latent space. 
\begin{align}
&Enc(t_i;\theta_{enc}) = f(Wx_i\: +\: Vy_i\: +\: b)
\end{align}
where, $\theta_{enc}=[W,\: V,\: b]$ are hidden layer learning parameters for $Enc$ and $f$ is a non-linear activation function.
The $Dec$ network takes $Enc(t_i;\theta_{enc})$ as an input and regenerates $t_i$ i.e., $t_{i}^{'}$.
\begin{align}
&Dec(Enc(t_{i};\theta_{enc});\theta_{dec}) = g([W'Enc(t)\:, \: V'Enc(t)]\: +\: b'),
\end{align}
\begin{align}
&t_{i}^{'} = Dec(Enc(t_{i};\theta_{enc});\theta_{dec}).
\label{equation:decoder_output}
\end{align}
Similarly, based on equations \eqref{equation:t_x_y} and \eqref{equation:decoder_output}, 
\begin{align}
&x_{i}^{'} = Dec(Enc([x_{i}, 0];\theta_{enc});\theta_{dec})
\label{equation:decoder_output_x}
\end{align}
\begin{align}
&y_{i}^{'} = Dec(Enc([0, y_{i}];\theta_{enc});\theta_{dec})
\label{equation:decoder_output_y}
\end{align}
where, $\theta_{dec}=[W',\: V',\: b']$ are hidden layer leaning parameters for $Dec$, and $g$ is non-linear activation function.
Subsequently, we define loss function, $\mathcal{J}_{T}(\theta)$, of the CorrNN as:
% loss function for binary data which autoencoder optimizes
\begin{align}
&\mathcal{J}_{T}(\theta) = \frac{1}{m}\sum_{i=1}^{m}(L(t_{i},t_{i}')\: +\: L(t_{i},x_{i}')\: +\: L(t_{i},y_{i}'))\: -\:  corr(Enc(x), Enc(y)).
\label{equation:ae_loss}
\end{align}

$\mathcal{J}_{T}(\theta)$ minimizes self-reconstruction error and cross-reconstruction error. It also maximizes correlation between the hidden representations of both the halves of the input data.
\begin{align}
&corr(Enc(x),Enc(y)) = \frac{\sum_{1}^{m}(Enc(x_{i})-\overline{Enc(x)})(Enc(y_{i})-\overline{Enc(y)})}{\sqrt{\sum_{1}^{m}(Enc(x_{i})-\overline{Enc(x)})^{2}\sum_{1}^{m}(Enc(y_{i})-\overline{Enc(y)})^{2}}}
\label{equation:correlation_hidden_space}
\end{align}

$L$ is the function for calculating reconstruction error; $m$ is the number of training samples in a batch; $Enc(x_i)$ and $Enc(y_i)$ are hidden layer representation of $x_i$ and $y_i$ respectively; $\overline{Enc(x)}$  and $\overline{Enc(y)}$ are mean vectors of every $Enc(x_i)$ and $Enc(y_i)$ respectively.

The $Dec$ is then used to map the continuous output of $G(z;\theta_{g})$ to discrete output $Dec(G(z;\theta_{g});\theta_{dec})$, where $z$ is the low dimensional random noise vector and $\theta_{g}$ are the learning parameters of G. $D( \cdot ; \theta_d)$ then predicts whether the generated discrete output is real or synthetic, where $\cdot$ can be any input and $\theta_d$ are the learning parameters of $D$.

\begin{figure}[H]
	\centering
	\includegraphics[scale=0.5]{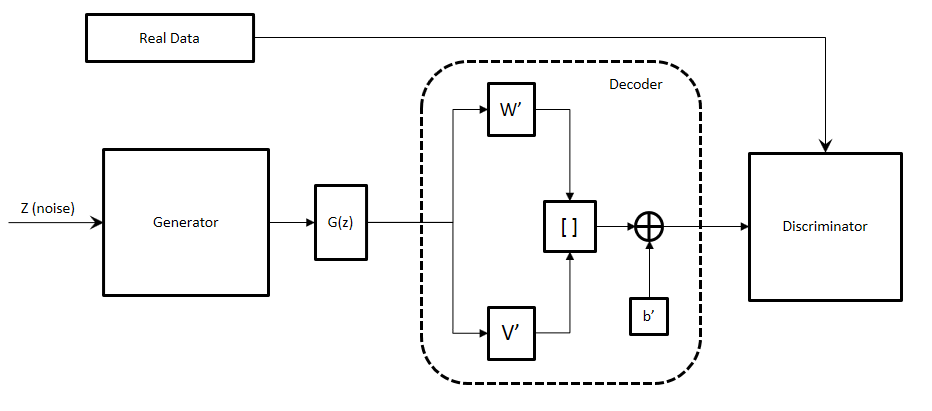}
	\caption{Block diagram GAN training module.}
	\label{fig:D_train}
\end{figure}
The objective function for the GAN model is:
% objective function for conditional GAN
\begin{align}
\min\limits_{G}\: \max\limits_{D}\: V(G,D)\: =\: {\mathbb{E}}_{t\sim p_{data}(t)}[\log\: D(t)] + {\mathbb{E}}_{z\sim p_{z}(z)}[\log(1-D(t_{z}))]
\end{align}
The update functions for $G$ and $D$ are as follows:
% gradient updates for generator and discriminator
\begin{align}
&\theta_{d}\: \leftarrow\: \theta_{d}\: +\: \alpha\nabla_{\theta_{d}}\frac{1}{m}\: \sum_{i=1}^{m}\log D(t_{i})\: +\: \log(1-D(t_{z_{i}}))\\
&\theta_{g,dec}\: \leftarrow\: \theta_{g,dec}\: +\: \alpha\nabla_{\theta_{g,dec}}\frac{1}{m}\: \sum_{i=1}^{m}\log D(t_{z_{i}}) 
\end{align}
where~~$t_{z_{i}} = Dec(G(z_{i}))$, $\theta_{d}$  and $\theta_{g}$ are learning parameters for $D$ and $G$ respectively, $\alpha$ is learning rate, $m$ is the number of training samples in the mini-batch, $t_{i}$ is original training input sample, and $t_{z}^{i}$ is synthetic sample generated with help of $G$ and $Dec$. 

To generate conditional discrete data, we follow the technique in \cite{MO14} i.e., append the apriori information $y$ to the noise prior $z$, hence $t_{z} = Dec(G([z,y]))$.
\\~\\
% Algorithm for conditional CorrGAN
\begin{algorithm}[H]
\SetAlgoLined
\DontPrintSemicolon
 $\theta_{d},\theta_{g},\theta_{enc},\theta_{dec}\: \leftarrow $ Initialize with random values.\;
 $\alpha\: \leftarrow $ learning rate.\;
 $m\: \leftarrow $ number of training samples.\;
 \While{pre-training epochs}{
 	Randomly sample $(x_{1},y_{1}), (x_{2},y_{2}), \ldots ,(x_{m},y_{m})$.\;
    Update $\theta_{enc}, \theta_{dec}$ by minimizing Eq. \eqref{equation:ae_loss} where $t_{i}=[x_{i},y_{i}]$.\;
 }
 \While{training epochs}{ 
  \For{d-steps}{
   //Update discriminator:\\
  	Randomly sample $z_{1}, z_{2},
  	\ldots ,z_{m} \in p_{z}$.\;
    Randomly sample $t_{1} = [x_{1},y_{1}], t_{2} = [x_{2},y_{2}],\ldots,t_{m} = [x_{m},y_{m}]$.\;
    $t_{z_{i}}\: \leftarrow\: Dec(G([z_{i}, y_{i}]))$.\;
    $\bar{t}_{z}\: \leftarrow\: \frac{1}{m}\sum_{1}^{m}\: t_{z_{i}}$.\;
    $\bar{t}\: \leftarrow\: \frac{1}{m}\sum_{1}^{m}\: t_{i}$.\;
    $\theta_{d}\: \leftarrow\: \theta_{d}\: +\: \alpha\Delta_{\theta_{d}}\frac{1}{m}\: \sum_{1}^{m}\log D(t_{i}, \bar{t})\: +\: \log(1-D(t_{z_{i}},\bar{t}_{z})).$
  }
  //Update generator and decoder:\;
  Randomly sample $z_{1}, z_{2},\ldots,z_{m} \in p_{z}$.\;
  Randomly sample $y_{1},y_{2},\ldots,y_{m}$ from training data.\;
  $t_{z_{i}}\: \leftarrow\: Dec(G([z_{i},y_{i}]))$.\;
  $\bar{t}_{z}\: \leftarrow\: \frac{1}{m}\sum_{1}^{m}\: t_{z_{i}}$.\;
  $\theta_{g,dec}\: \leftarrow\: \theta_{g,dec}\: +\: \alpha\Delta_{\theta_{g,dec}}\frac{1}{m}\: \sum_{1}^{m}\log D(t_{z_{i}},\bar{t}_{z})$.
 }
 \caption{Algorithm for conditional CorrGAN}\label{algo:skill-GAN}
\end{algorithm}

The training process of CorrGAN is demonstrated in Algorithm \ref*{algo:skill-GAN}. During training, the first step is pre-training of the CorrNN, where it minimizes $\mathcal{J_T(\theta)}$ according to Equation \ref*{equation:ae_loss}. The second step is training $G$ and $D$, where we use mini-batch averaging and eventually, $G$ and $D$ converge.

\section{Experimental Setting and Results}\label{section:ExpAndResults}
In this section, we present results of Algorithm \ref*{algo:skill-GAN}. The experiments were carried out on a machine with configuration (Intel i7-4770, 8GB RAM, Intel HD 4600, Ubuntu 64bit) environment.
% First, we discuss the data pre-processing, carried out to prepare first dataset for training GAN. 

\subsection{Results}\label{subsection:ExpAndResults:Results}
\begin{figure}[H]
\centering
\begin{tabular}{|c|c|c|c|c|}
\hline\small
\subfloat{\includegraphics[scale=\mnistImagesWidth]{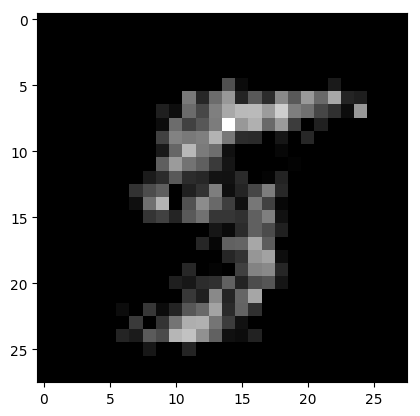}} &
\subfloat{\includegraphics[scale=\mnistImagesWidth]{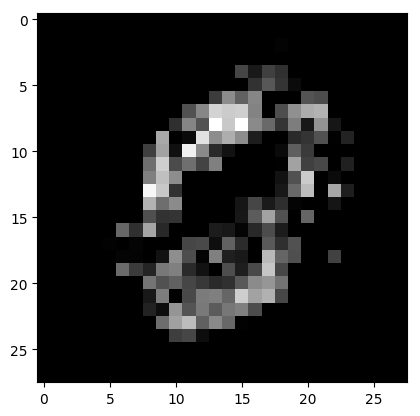}} &
\subfloat{\includegraphics[scale=\mnistImagesWidth]{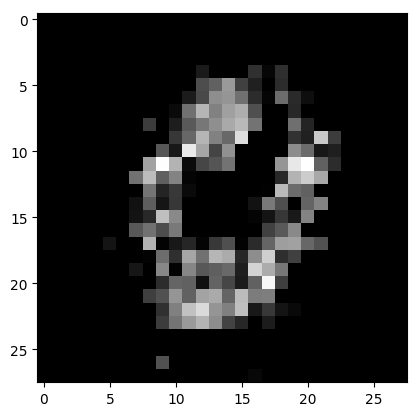}} &
\subfloat{\includegraphics[scale=\mnistImagesWidth]{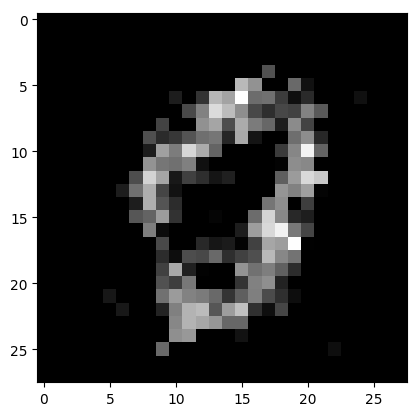}} &
\subfloat{\includegraphics[scale=\mnistImagesWidth]{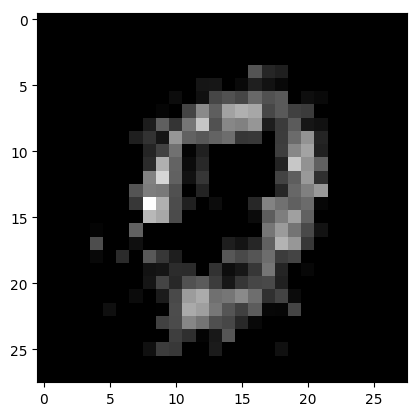}} \\
\hline
\subfloat{\includegraphics[scale=\mnistImagesWidth]{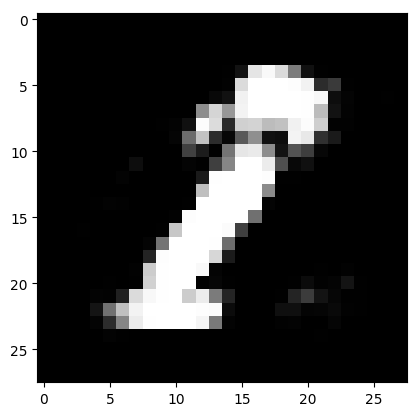}} &
\subfloat{\includegraphics[scale=\mnistImagesWidth]{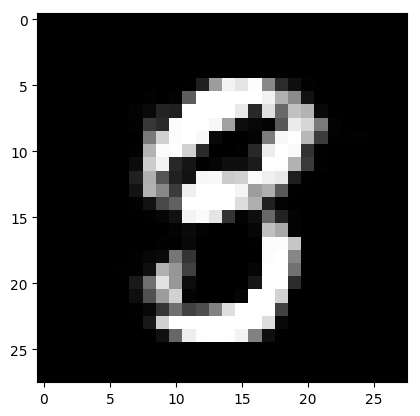}} &
\subfloat{\includegraphics[scale=\mnistImagesWidth]{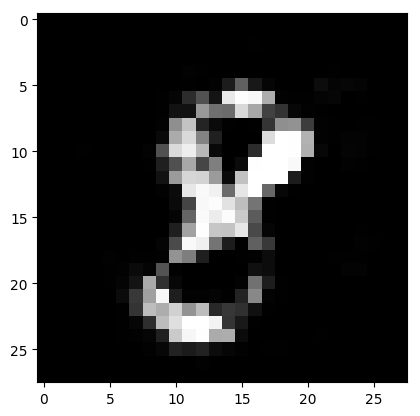}} &
\subfloat{\includegraphics[scale=\mnistImagesWidth]{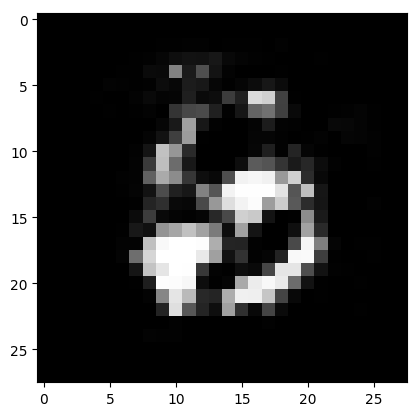}} &
\subfloat{\includegraphics[scale=\mnistImagesWidth]{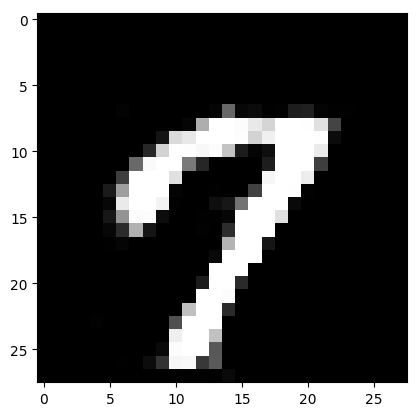}} \\
\hline
\scriptsize{100 epochs}&
\scriptsize{400 epochs}&
\scriptsize{700 epochs}&
\scriptsize{1000 epochs}&
\scriptsize{1300 epochs}\\
\hline
\end{tabular}
\caption{Generated images by medGAN(top row) and CorrGAN(bottom row) for given epochs}
\label{fig:mnist_generated_comparison}
\end{figure}

 As mentioned earlier, we validate our model on two datasets- MNIST \cite{MNIST10} and skill dataset. We choose MNIST due to its conformity as a standard dataset for machine learning applications and it provides an inherent correlation which suits our application. The images generated are not conditioned and are generated by giving noise as the input to the generator. The experiment aims to warrant the usage of CorrNN instead of Vanilla Autoencoder for translation of discrete data into continuous space. Figure \ref*{fig:mnist_generated_comparison} shows the comparative results between medGAN \cite{CBM17} (which uses Vanilla Autoencoder) and CorrGAN for MNIST dataset on different epochs.  
%   The experiments were carried out on a machine with configuration (Intel i7-4770, 8GB RAM, Intel HD 4600, Ubuntu 64bit) environment. We perform our experiments on the above mentioned skill dataset and the standard MNIST dataset \cite{MNIST10} to confirm the generality. Table \ref{table:exp_setup} gives the details pertaining to the skill data at hand. Figure \ref{fig:mnist_generated_100_epoch},  \ref{fig:mnist_generated_700_epoch} and \ref{fig:mnist_generated_1400_epoch} shows the comparitive results between medGAN \cite{CBM17} and CorrGAN for MNIST dataset on different epochs.  

\begin{figure}[H]
 	\centering
 	\begin{tabular}{|c|c|c|c|c|c|}
 		\hline\small
 		\subfloat{\includegraphics[scale=\mnistImagesWidth]{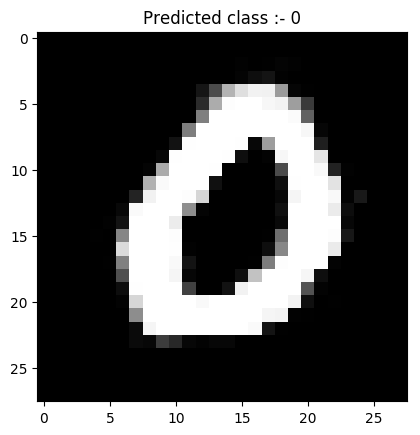}} &
 		\subfloat{\includegraphics[scale=\mnistImagesWidth]{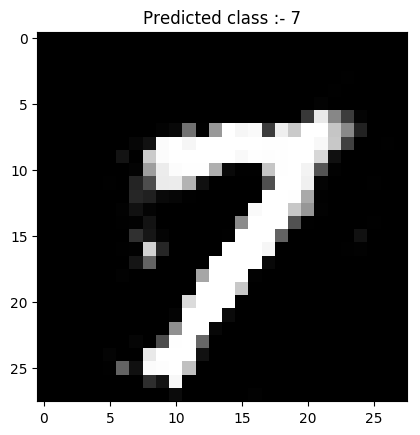}} &
 		\subfloat{\includegraphics[scale=\mnistImagesWidth]{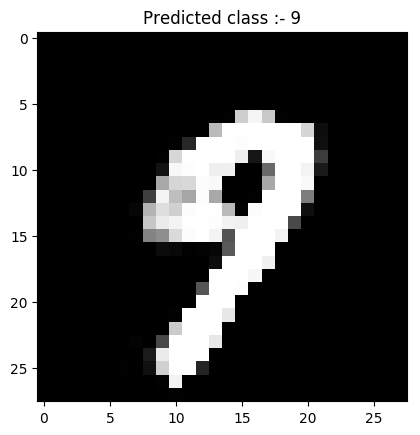}} &
 		\subfloat{\includegraphics[scale=\mnistImagesWidth]{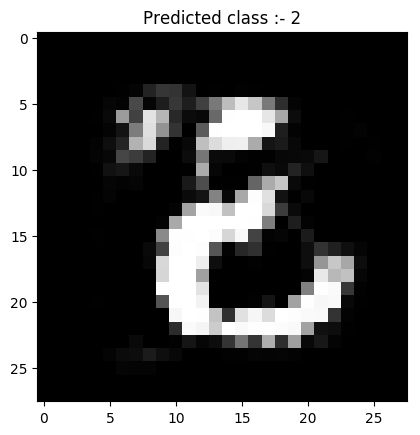}} &
 		\subfloat{\includegraphics[scale=\mnistImagesWidth]{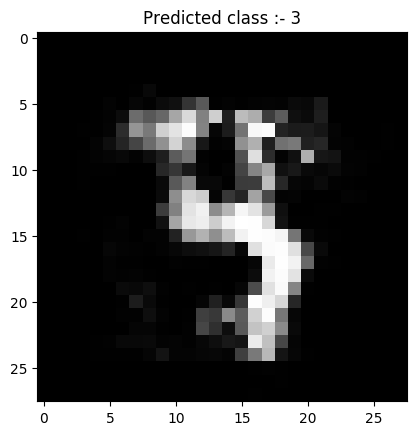}} &
 		\subfloat{\includegraphics[scale=\mnistImagesWidth]{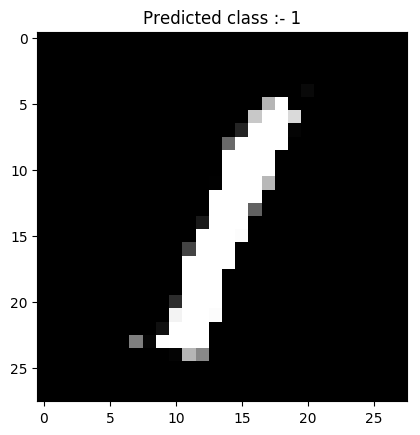}} \\
 		\multirow{ 2}{*}{}\scriptsize{Predicted} & \scriptsize{Predicted} &
 		\scriptsize{Predicted} & \scriptsize{Predicted} & \scriptsize{Predicted}  & \scriptsize{Predicted} \\
 		\scriptsize{label: 0} & \scriptsize{label: 7} & \scriptsize{label: 9} & \scriptsize{label: 2} &\scriptsize{label: 3} & \scriptsize{label: 1} \\
 		\hline
 		\subfloat{\includegraphics[scale=\mnistImagesWidth]{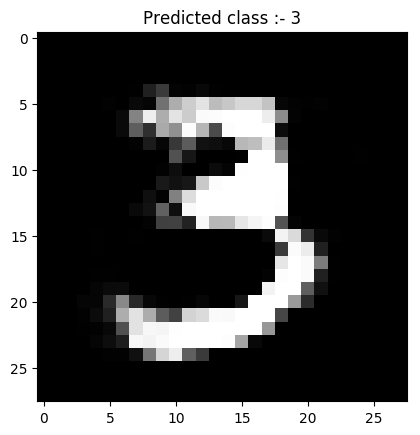}} &
 		\subfloat{\includegraphics[scale=\mnistImagesWidth]{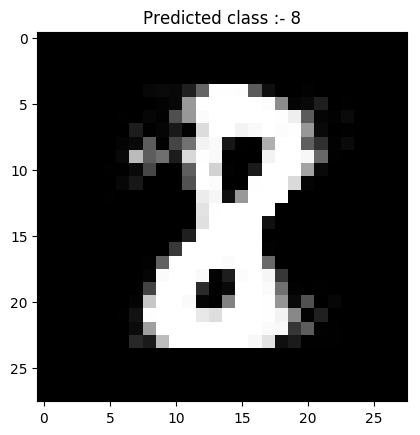}} & \subfloat{\includegraphics[scale=\mnistImagesWidth]{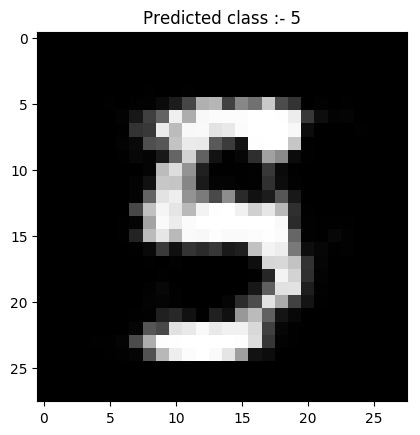}} &
 		\subfloat{\includegraphics[scale=\mnistImagesWidth]{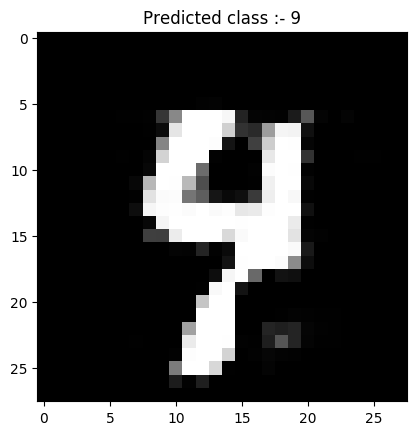}} &
 		\subfloat{\includegraphics[scale=\mnistImagesWidth]{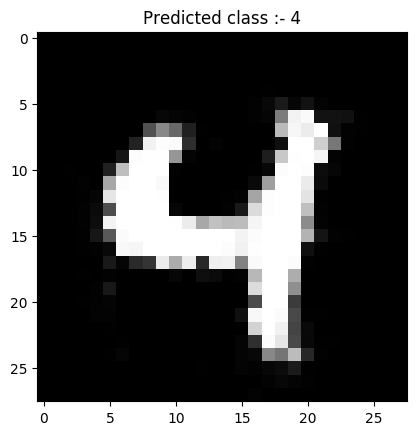}} &
 		\subfloat{\includegraphics[scale=\mnistImagesWidth]{8}} \\
 		\multirow{ 2}{*}{}\scriptsize{Predicted} & \scriptsize{Predicted} &
 		\scriptsize{Predicted} & \scriptsize{Predicted} & \scriptsize{Predicted}  & \scriptsize{Predicted}  \\
 		\scriptsize{label: 3} & \scriptsize{label: 8} & \scriptsize{label: 5} & \scriptsize{label: 9} &\scriptsize{label: 4} & \scriptsize{label: 8} \\
 		\hline
 	\end{tabular}
 	\caption{CorrGAN-MNIST samples and predicted classes}
 	\label{fig:mnist_generated_prediction}
 \end{figure}

For generating conditional data on the MNIST dataset, we exploit the inherent correlation between pixels of the image of a certain number. We replace the top half of the MNIST images with noise and pass it as input to the generator. The generator then generates a latent space, given half of the actual image. This space is then passed to the Decoder which produces a complete image. Some resultant images are given in Figure \ref*{fig:mnist_generated_prediction}.
%We pass the images generated by CorrGAN through the scikit-learn's built-in SVM classifier \cite{SCI11}(this classifier has 92\% accuracy in classifying the MNIST dataset).
 
% \begin{figure}[H]
%  \begin{tabular}{ccccc}
%  \subfloat{\includegraphics[width = 0.18\textwidth]{0}} &
%  \subfloat{\includegraphics[width = 0.18\textwidth]{7}} &
%  \subfloat{\includegraphics[width = 0.18\textwidth]{9}} &
%  \subfloat{\includegraphics[width = 0.18\textwidth]{2}} &
%  \subfloat{\includegraphics[width = 0.18\textwidth]{1}} \\
%  \subfloat{\includegraphics[width = 0.18\textwidth]{3}} &
%  \subfloat{\includegraphics[width = 0.18\textwidth]{6}} &
%  \subfloat{\includegraphics[width = 0.18\textwidth]{5}} &
%  \subfloat{\includegraphics[width = 0.18\textwidth]{4}} &
%  \subfloat{\includegraphics[width = 0.18\textwidth]{8}} \\
%  \subfloat{\includegraphics[width = 0.18\textwidth]{2_2}} &
%  \subfloat{\includegraphics[width = 0.18\textwidth]{7_7}} &
%  \subfloat{\includegraphics[width = 0.18\textwidth]{0_0}} &
%  \subfloat{\includegraphics[width = 0.18\textwidth]{2_other}} &
%  \subfloat{\includegraphics[width = 0.18\textwidth]{9_9}} \\
%  \subfloat{\includegraphics[width = 0.18\textwidth]{1_1}} &
%  \subfloat{\includegraphics[width = 0.18\textwidth]{3_3}} &
%  \subfloat{\includegraphics[width = 0.18\textwidth]{480}} &
%  \subfloat{\includegraphics[width = 0.18\textwidth]{3_other}} &
%  \subfloat{\includegraphics[width = 0.18\textwidth]{4_4}} \\
%  \end{tabular}
%  \caption{Images generated by CorrGAN and their prediction by SVM Classifier}
%  \label{fig:mnist_generated_prediction}

% \end{figure}  

We next perform the experiment on candidate skill dataset. Due to scarcity of clean data, appropriate data pre-processing was undertaken as detailed in Appendix  \ref*{appendix:datapreprocessing}. Table \ref*{table:exp_setup} gives the details pertaining to the skill data at hand.
\begin{table}[H]
	\centering
	\begin{tabular}{|l|c||l|c|} 
		\hline
		Description &  Statistics & Description &  Statistics\\
		\hline\hline
		Total different Job profiles  & 6 & Max number of skills of a candidates& 108\\ 
		\hline
		Total candidates & 9686 & Min number of skills of a candidates& 0\\
		\hline
		Total skills  & 7055 & Median number of skills  & 5\\
		\hline
		Avg number of profiles per job & 1614 & Training Time& 15Hrs.\\
		\hline
		Max number of profiles of a job & 2248 & Pre-training epoch & 150\\
		\hline
		Min number of profiles of a job & 460 & Epoch &1000\\
		\hline
		Avg number of skills per candidates& 6 & Batch size & 100 \\
		\hline
	\end{tabular}
	\caption{Training data statistics for input skill-set data, after data preprocessing}
	\label{table:exp_setup}
\end{table}
%\addfigureHW{apna}{6cm}{1cm}{The graphs capture the probabilities of occurrence of each skill in feature (training) set versus that in the generated set, for the two models (medGAN, CorrGAN). Each point corresponds to a skill in our database. Ideally, in order to replicate the characteristic of the feature set, the points should fall on the line \textit{y=x} implying equal probability in either set.}	\label{fig:Scatter}

We generate results conditioned on a given profession. The results for each profile is given in Table \ref*{table:snippet_of_generated_data_CorrGAN}.
\begin{table}[H]
\centering
{\renewcommand\arraystretch{1.25}
\begin{tabular}{|l|l|l|} \hline
A prior input: Profession & \multicolumn{2}{p{7cm}|}{Generated Skills}\\
\hline \hline
\multirow{3}{*}{Web designer}& \multicolumn{2}{p{7cm}|}{ms sql server, oracle, database, sql, java, .net} \\\cline{2-3}
& \multicolumn{2}{p{7cm}|}{mysql, javascript, c++, jira, html} \\\cline{2-3}
& \multicolumn{2}{p{7cm}|}{javascript, android, php, jquery, java} \\ \hline

\multirow{3}{*}{Net developer}& \multicolumn{2}{p{7cm}|}{xml, illustrator, css, html} \\\cline{2-3}
& \multicolumn{2}{p{7cm}|}{graphic design, illustrator, photoshop, html} \\\cline{2-3}
& \multicolumn{2}{p{7cm}|}{hibernate, xml, spring, jdbc, servlets, jquery, struts, ajax, j2ee, servlet, html, jsf, jsp, java} \\ \hline

\multirow{3}{*}{Java developer}& \multicolumn{1}{p{7cm}|}{oracle, eclipse, jdbc, j2ee, sql, jsp, java} \\\cline{2-3}
& \multicolumn{1}{p{7cm}|}{ javascript, asp.net, c\#, asp, jquery, ms asp, .net} \\\cline{2-3}
& \multicolumn{2}{p{7cm}|}{javascript, j2ee, css, html, jsp, java} \\ \hline

\multirow{3}{*}{Application developer}& \multicolumn{1}{p{7cm}|}{asp.net, c\#, asp, ms asp, .net} \\\cline{2-3}

& \multicolumn{1}{p{7cm}|}{unix, mysql, wordpress, vb.net, programming, xml, rest, javascript, linq, scrum, python, soap, c\#, php, css3, eclipse, visual studio, ibm rad, sublime, scss, jquery, linux, web development, agile, ajax, pl/sql, iis, jira, json, web design, git, sql, angular js, html, windows, java} \\\cline{2-3}

& \multicolumn{1}{p{7cm}|}{mysql, xml, ms sql server, javascript, php, sql server, c++, json, sql, html, java} \\ \hline

Application support analyst& \multicolumn{2}{p{7cm}|}{asp.net, c\#, asp, database, ms asp, iis, sql, .net} \\ \hline

\end{tabular}} %\vspace{0.1in}
\caption{Snippet of Generated data for a given profession using CorrGAN.}
\label{table:snippet_of_generated_data_CorrGAN}
\end{table}

%The length of the skill vectors are more. We see that the matrix of number of profiles versus vector length would be shorter and narrower in our cases. This dearth of samples and increased number of skills would lead to insufficient training, hence implying the requirement of more training data. 
%\subsection{Predictions of Generated Images}
%We train the scikit-learn built-in SVM on the 
 %\textbf{This model finds its application in the form of an AI career counselling agent which can suggest skills to job-seeking candidates that are present in the generated samples corresponding to the candidate's profession but are lacking in their profile. }
 As a simple use case of synthetic results for a career counseling AI agent, we can consider a candidate's current profession and compute a set difference between his/her current skills and identify skills from synthetic data to suggest to the candidate as skills that they should acquire for career progression.
 Note that we have tested the model for two given data sets, i.e, user skill vectors and MNIST. However, this model can be used for inherently correlated data.
 
 \subsection{Evaluation Metrics}\label{subsection:ExpAndResults:EvaluationMetrics}
  To evaluate the generated images by our model we pass the images generated by CorrGAN through the scikit-learn's built-in SVM classifier \cite{SCI11} which is known to have 92\% accuracy in classifying the MNIST dataset. For the skill dataset, we use scatter plots to compare probabilities of occurrence of skills in the original and generated (synthetic) data. We calculate the mean-squared error of scatter points with respect to the line \textit{y=x} to outline a comparison between CorrGAN and medGAN. Furthermore, we define a correlation metric to evaluate the joint occurrences of skills in the generated data as compared to that of the original data. 
%% should add 7 more lines, to get the next content onto the next page.

We find 69\% of SVM labeled images to be matching our perceptual estimates\footnote{Majority vote by authors on 100 SVM samples}. Some of the results are shown in Figure \ref*{fig:mnist_generated_prediction}.

 \begin{figure}[H]
 	\centering
 	\includegraphics[height=6cm,width=13cm]{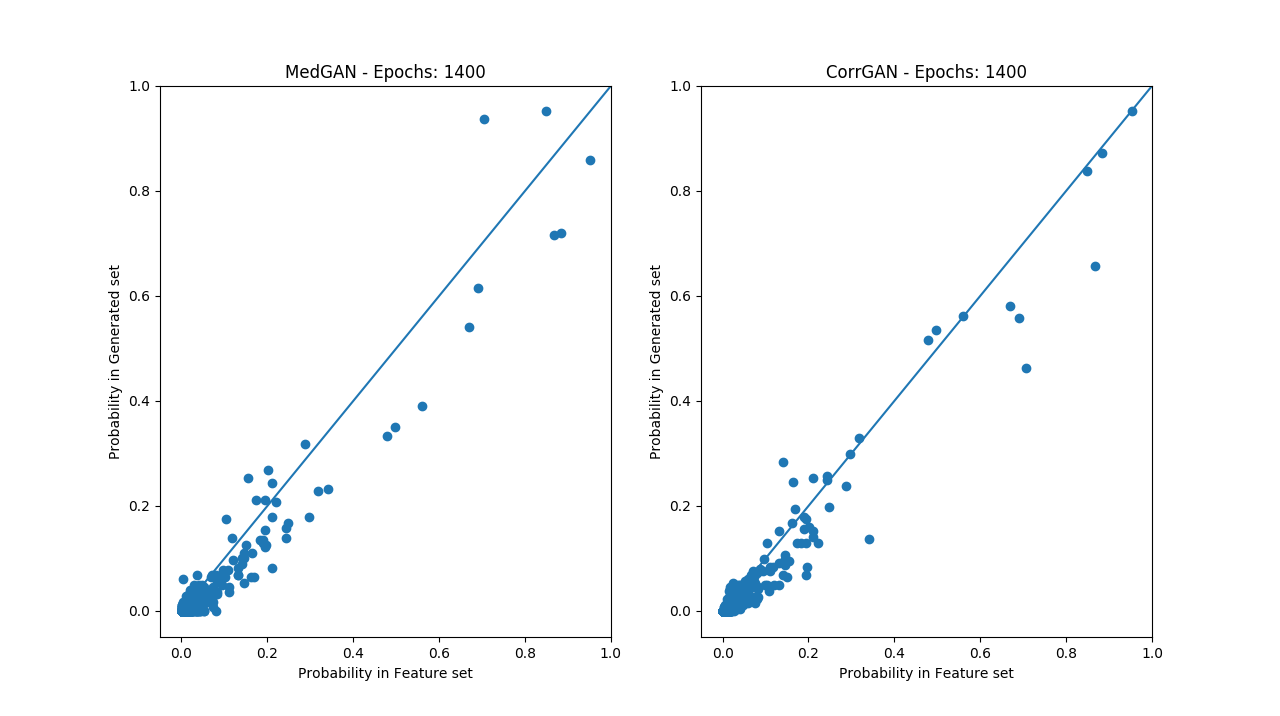}
 	\caption{The graphs capture the probabilities of occurrence of each skill in feature (training) set versus that in the generated set, for the two models (medGAN, CorrGAN). Each point corresponds to a skill in our database. Ideally, in order to replicate the characteristic of the feature set, the points should fall on the line \textit{y=x} implying equal probability in either set.}
 	%%	 The error between correlation matrix of medGAN and that of original data is calculated and subtracting it from 1 gives us the medGAN plot. Similarly,The error between correlation matrix of CorrGAN and that of original data is calculated and subtracting it from 1 gives us the CorrGAN plot}
 	\label{fig:Scatter}
 \end{figure}
 In Figure \ref*{fig:Scatter}, the probability measure of skills' occurrence in the input set to the generated set is plotted. The mean square-error(MSE) is calculated to account the variation of scatter points from the line \textit{y=x}. Consequently, the MSE estimate for medGAN is $7.567\times 10^{-5}$ and corrGAN is $5.049\times 10^{-5}$ (at 1400 epochs). However, Figure \ref*{fig:Scatter} demonstrates quantification of the occurrence of a skill, rather than considering correlation among two skill vectors, which motivates us to use a different evaluation metric, to quantify the joint occurrence of skills.

\begin{figure}[H]
  	\centering
  	\includegraphics[width=0.56\linewidth]{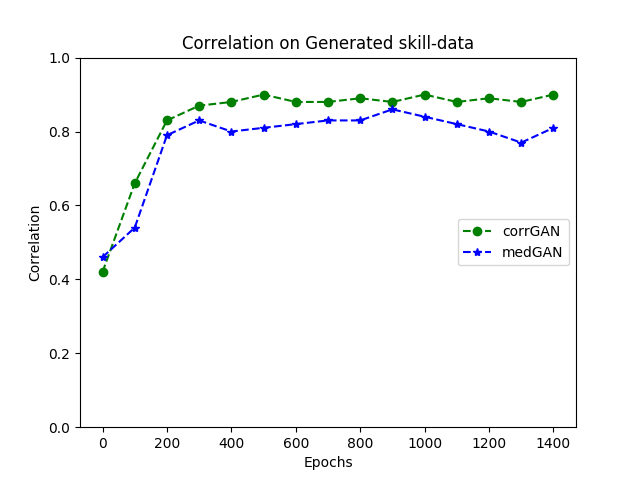}
  	\caption{Correlation of generated data vs epochs. Note that we calculate the correlation matrix for medGAN, CorrGAN and original data by Algorithm \ref*{algo:corMatrix_Algorithms}.}
  	%	 The error between correlation matrix of medGAN and that of original data is calculated and subtracting it from 1 gives us the medGAN plot. Similarly,The error between correlation matrix of CorrGAN and that of original data is calculated and subtracting it from 1 gives us the CorrGAN plot}
  	\label{fig:SKILL_Correlation_On_Generated}
  \end{figure}

 %Figure \ref{fig:Scatter} shows scatter plots  probability of skill occurrence in original and generated data. \\
 Hence, to measure the skill coherence, we calculate the correlation between the generated skills for the medGAN and CorrGAN models for every 100 epochs,  by computing the cooccurrence of two skills. Cooccurrence matrices are calculated for both the original data and the generated data by Algorithm \ref*{algo:corMatrix_Algorithms} using respective data matrix. Note that the matrices are symmetric and are \textit{skill vs skill} matrix with every cell as the intersection of row skill and column skill, being the total co-occurrences of the two skills. The matrices are then normalized so that each value lies in [0,1]. The original data's co-occurrence matrix is subtracted from that of the generated data to calculate the error matrix. The error mean is taken to reduce the matrix to a scalar.  We plot the correlation in Figure \ref*{fig:SKILL_Correlation_On_Generated}. We infer from the plot, that correlation improves in our model over the number of epochs.  
 
 % Algorithm for conditional CorrGAN
 \begin{algorithm}[h!]
 	\SetAlgoLined
 	\DontPrintSemicolon
 	Let $vectors\: \leftarrow$ no of samples/profiles\;
  	Let $dim\: \leftarrow$ no of skills\;
  	$inputMatrix[vectors][dim]\: \leftarrow$ Data Matrix\;
  	%$corMatrix[dim][dim]\: \leftarrow $ \text{\large O} .\;
  	Initialize $corMatrix[dim][dim]\: \leftarrow $ Zero Matrix\;
  	Initialize threshold $\alpha\: \leftarrow $ Preset value $\in$ [0, 1]\;
  	%$threshold\: \leftarrow $ Initialize with pre-defined value $\in$ [0,1].\;
  	\For{i=1:dim-1}{
  		\For{j=i+1:dim}{
  			\For{k=1:vectors}{
  				\eIf{inputMatrix[k][i]$>\alpha$ and inputMatrix[k][j]$>\alpha$}{
  					corMatrix[i][j]+=1.\;
  				}{}
  			}
  			corMatrix[j][i]:=corMatrix[i][j].\;
  		}
  	}
  	corMatrix:=corMatrix/vectors.\;
  	\caption{Build correlation matrix of given skill vector}
  	\label{algo:corMatrix_Algorithms}
 \end{algorithm}

\section{Future work}\label{section:FutureWork}
  Candidate data is mostly proprietary content and thus its availability is scarce. The dataset consisted of 24,934 candidate profiles, out of which 6,762 profiles had an empty skill section and hence had to be removed from the dataset. Thus, a module which can extract skills from other sections of profiles like projects and experience can be developed.
% Also, the available data is in non-standard formats and requires pre-processing for cleaning and organizing which further reduces the dataset size as every sample is not fit for training. 
Further, certain imperfections in cleaning the data - creating valid skill tokens from text in skill section - affect the experiments. An interesting future improvement can be to map the skills to a predefined dictionary of skill by calculating the edit distance between them or using a more robust algorithm such as Fuzzy Matching \cite{CSGK03}.

%Training Deep Neural Networks or GAN would require sufficiently large dataset. For the same, testing this model on a larger dataset would be an interesting exercise. 
As a non-trivial extension, our model can be extended to condition on other parameters, like geographical regions, education background, organization, in a candidate profile.
% This provided we have significant amount of training samples in each case.

\section{Acknowledgement}\label{section:Acknowledgement}
  The authors of this work would like to thank HireValley Pvt. Ltd. for providing candidate profile data, which was pivotal to the experiments conducted. 
%\bibliographystyle{ieeetr}
%\bibliography{CorrGAN_Bibliography}

\appendix{}
\section{Data Pre-Processing}
\label{appendix:datapreprocessing}
We obtained 24,934 candidate profiles in JSON format.
A typical candidate profile contains work experience, educational qualification, projects, additional information apart from skill data. However, the experiments presented in this work only require use of skill data and the candidate's current profession. An example of input data containing skills as discrete tokens and profession is given in Table \ref*{table:input_snippet}. 

It must be noted that 8486 candidate profiles (out of 24934) contained narrative in the skill section. Table \ref*{table:input_discarded} lists one such profile. 
Deriving skills from such narrative requires specific data mining approaches.
We also note that, as far as skills in the IT sector are concerned, most skills are of length 15 characters or less\footnote{ The average length of skills in our data is 11.4173 characters with a standard deviation of $\pm$ 4.7876. Hence, we infer that our threshold of 15 is suitable for good results.}. Thus, in this work we do not focus on data points with comprehension in skill section and ignore profiles with skill lengths exceeding 15 characters. Note that the threshold 15 is an empirically derived constant, and can be changed without affecting performance of the algorithm presented. 
%Errors pertaning to improper placement of delimiters, addition of punctuations and spelling errors still.  All Unicode words are removed.
\begin{table}[H]
	\centering
	\begin{tabular}{|l|p{9cm}|} 
		\hline
		Profession & Skills\\
		\hline\hline
		Java Developer & Java, J2EE, Servlets, Jsp, JQuery, Spring 2.5, Spring MVC\\ 
		\hline
		Application Developer & .NET, SQL, ASP .Net, VB.NET, C\#, Oracle, WCF \\ 
		\hline
		Applications Engineer & Java, J2EE, Servlets, Jsp, JQuery, Spring 2.5, Spring MVC, SOAP \\ 
		\hline
		Application Support Analyst & UNIX, AIX, Solaris, Sun Storage, Sun SPARC, Sun Ultra, HP, VNC \\ 
		\hline
		Net Developer & C\#, SQL, ASP, ASP.NET, MS ASP, MS SQL SERVER, SQL SERVER \\ 
		\hline
        Java Developer & JAVA, JAVASCRIPT, JSP, JUNIT, HTML, SOAP, XML \\ 
		\hline
	\end{tabular}
	\caption{Snippet of Input data}
	\label{table:input_snippet}
\end{table}

\begin{table}[h]
\centering
\begin{tabular}{| p{3cm} | p{10cm} |} 
\hline
Net Developer & Gathering and analysis of requirements and delivery of solutions (1 year). 
High experience level in computer repair, assemble and analysis of requirements for a specific computer (3 years). Operating Windows XP and newer versions, Installation and configuration (3 years), Management of the Linux Operating System, installation, basic configuration and installation of basic programs(Less than 1 year). Basic knowledge of Java, HTML, CSS, C\#, JavaScript. (Less than 1 year), Basic knowledge of MySQL DataBase (Less than 1 year), Basic management of Netbeans IDE, VS.NET, Photoshop, Microsoft Office (1 year), Basic knowledge of SQL Server. (1 year)\\
\hline
\end{tabular}
\caption{Candidate profile with comprehension in skill section. }
\label{table:input_discarded}
\end{table}
We maintain a superset of skills and professions called \textit{skillDictionary} and \textit{professionDictionary}, created by taking the union of all skills and professions in the input profession-skills pairs respectively. These pairs of profession-skills is converted to pairs of \textit{professionVector}-\textit{skillVector}. The  \textit{skillVector} is binary, containing ones and zeros for the skills present and absent respectively in a candidate profile, whereas \textit{professionVector} is one-hot, containing a $1$ in an appropriate position, corresponding to a candidate's current profession.
\end{document}